\documentclass[runningheads]{llncs} 

\usepackage{eccv}

\usepackage{eccvabbrv}

\usepackage{graphicx}
\graphicspath{{./}}
\usepackage{booktabs}
\usepackage{caption}
\usepackage{multirow}
\usepackage{microtype}  

\usepackage{iftex}
\ifPDFTeX
  \usepackage[accsupp]{axessibility}  
\fi

\usepackage{hyperref}

\usepackage{orcidlink}

\begin{document}

\title{DeCoFlow: Structural Decomposition of Normalizing Flows for Continual Anomaly Detection\thanks{Code is available at \url{https://github.com/crimama/DeCoFlow}.}}

\titlerunning{DeCoFlow: Continual Anomaly Detection via NF Decomposition}

\author{Hun Im\inst{1}\orcidlink{0009-0003-6056-1870} \and
	Jungi Lee\inst{1}\orcidlink{0000-0002-0496-5518} \and
	Subeen Cha\inst{1}\orcidlink{0009-0000-7175-9928} \and
	Pilsung Kang\inst{1}\orcidlink{0000-0001-7663-3937}}

\authorrunning{Im et al.}

\institute{Seoul National University, Seoul, Korea\\
	\email{\{hun\_im,jungi\_lee,subeen\_cha,pilsung\_kang\}@snu.ac.kr}}

\maketitle

\begin{abstract}
	In industrial environments, new product categories arrive sequentially, requiring continual anomaly detection without access to past data. Normalizing Flows (NFs) provide exact density estimation but suffer from catastrophic forgetting as parameter updates across tasks distort the density manifold. While parameter isolation can prevent interference, it must preserve the strict invertibility and Jacobian validity of NFs. To satisfy these requirements, we exploit the inherent property that affine coupling layers maintain transformation validity regardless of subnet parameterization. Based on this, we propose DeCoFlow, which decomposes subnets into a frozen universal base and task-specific low-rank adapters to isolate updates. We further introduce Task-Specific Alignment, Auxiliary Coupling Layers, and Tail-Aware Loss to compensate for frozen-base rigidity. DeCoFlow achieves state-of-the-art image-level AUROCs of 98.40\% on MVTec-AD and 93.00\% on VisA, while maintaining parameter-level zero forgetting (0.00\% FM under correct routing) with only 2.27M parameters per task.
\end{abstract}


\section{Introduction}
\label{sec:intro}

In industrial manufacturing, product lines evolve over time, requiring anomaly detectors to accommodate new classes sequentially. Recent anomaly detection has moved from one-class to multi-class settings~\cite{uniad, omnial}, yet most methods assume a static data distribution, whereas practical manufacturing requires learning tasks sequentially under storage and privacy constraints~\cite{delange2021clsurvey}, forcing models to learn without past data and leading to catastrophic forgetting~\cite{mccloskey1989, french1999}.

Normalizing Flows (NFs) have emerged as a powerful baseline because they provide exact density estimation and naturally avoid the identity mapping problem of reconstruction-based methods. However, extending NFs to continual learning is challenging as catastrophic forgetting is significantly more pronounced in density-estimation models. Unlike classification models that only need to preserve decision boundaries, NFs rely on maintaining the entire density manifold to score anomalies; even minor parameter interference during sequential updates can thus severely distort the learned density~\cite{dne}.

Because most existing CL methods~\cite{ewc, li2017lwf} were designed for discriminative tasks, applying them to NFs reveals a fundamental dilemma. Replay methods mitigate forgetting but introduce fidelity limitations that degrade the precise density estimation required by NFs~\cite{dgr}. Parameter-isolation methods prevent interference but lead to linear model growth~\cite{rusu2016progressive}. Although parameter-efficient fine-tuning (PEFT) techniques like LoRA offer a scalable alternative~\cite{gainlora, mingle}, applying them directly to NFs is constrained by the strict invertibility and exact Jacobian requirements of flow models~\cite{dinh2016realnvp}.

To address this, we exploit a structural property of coupling-based NFs: transformation validity is preserved regardless of subnet parameterization. Based on this insight, we propose DeCoFlow, a framework that decomposes each scale/shift subnet into a frozen universal base and learnable task-specific low-rank residuals~\cite{dinh2016realnvp}. This subnet-level decomposition preserves invertibility and Jacobian validity (\cref{sec:theoretical_foundation}) while enabling task-wise adapter isolation. To compensate for frozen-base rigidity, we introduce Task-Specific Alignment (TSA), Auxiliary Coupling Layers (ACL), and Tail-Aware Loss (TAL) (\cref{sec:method}). Our main contributions are:
\begin{itemize}
	\item \textbf{Structural Design Principle:} We establish that subnet independence in coupling-based NFs enables parameter isolation for continual learning, empirically confirming this architectural specificity through cross-architecture comparisons.
	\item \textbf{DeCoFlow Framework:} We propose DeCoFlow, decomposing each coupling subnet into a frozen universal base and task-specific low-rank adapters, achieving parameter-level zero forgetting under correct routing with 2.5\% (2.27M) overhead per task.
	\item \textbf{Frozen-Regime Compensators:} To mitigate frozen-base rigidity, we introduce TSA for input alignment, ACL for residual correction, and TAL for capacity redistribution, validating their synergy through factorial and block-wise analysis.
\end{itemize}


\section{Related Work}
\label{sec:related_work}

\subsection{Deep Anomaly Detection}
Recent anomaly detection has transitioned to the Unified AD paradigm, covering multiple classes with a single model~\cite{uniad, omnial}. However, existing reconstruction-based methods are vulnerable to the identity mapping problem, where even anomalous data is reconstructed. In contrast, Normalizing Flows (NFs) structurally avoid this through direct likelihood optimization and have demonstrated strong performance~\cite{fastflow, msflow}. While recent methods introduce GMMs~\cite{hgad} or vector quantization~\cite{vqflow} for multi-class modeling, their fixed capacity limits adaptation to evolving data. Continual learning is thus essential for dynamic industrial environments.

\subsection{Continual Learning}
Continual learning methods for mitigating catastrophic forgetting~\cite{mccloskey1989, french1999} fall into three categories: regularization (EWC~\cite{ewc}, MAS~\cite{mas}), replay (GEM~\cite{gem}, DGR~\cite{dgr}), and architecture-based isolation including neuron masking~\cite{mallya2018packnet, serra2018hat} and PEFT such as LoRA~\cite{lora}. Among these, LoRA-based methods have gained prominence for structurally preventing parameter interference. Recent extensions include MoE-based routing (GainLoRA~\cite{gainlora}, MINGLE~\cite{mingle}), geometric constraints (AnaCP~\cite{anacp}, CaLoRA~\cite{calora}), and dynamic subspace allocation (CoSO~\cite{coso}). However, these techniques target discriminative decision boundaries, and cannot be directly applied to density-based models without risking likelihood manifold collapse.

\subsection{Continual Learning in Anomaly Detection}
\label{sec:continual_ad}

Continual anomaly detection methods extend the aforementioned CL paradigms. \textbf{Replay}-based approaches mitigate forgetting by retaining past data or generating synthetic samples, utilizing techniques like incremental coresets~\cite{cadic} or diffusion replay~\cite{replaycad}. However, their effectiveness is strictly bounded by memory budgets and replay fidelity. Meanwhile, \textbf{constraint-driven} methods~\cite{dne, cfrdc} stabilize model updates to reduce forgetting, but inevitably suffer from the inherent stability-plasticity dilemma when learning new tasks. \textbf{Dynamic architecture} methods mitigate interference through structural changes. Parameter-isolation designs~\cite{surprisenet, rusu2016progressive} scale linearly with the number of tasks, whereas prompt-based variants~\cite{mtrmb, ucad} remain overly sensitive to prompt quality. None of these approaches exploits the inherent structural properties of NFs for safe parameter isolation.


\section{Structural Basis for Parameter Isolation}
\label{sec:theoretical_foundation}

The structural basis for parameter isolation in our framework lies in the architectural independence of affine coupling layers. We exploit this property to decompose the subnets within Normalizing Flows (NFs). This approach is rooted in the mathematical guarantee that the coupling mechanism preserves invertibility and exact Jacobian computation regardless of its internal subnet architecture~\cite{dinh2016realnvp}.

The key rationale is that coupling layer invertibility is guaranteed by the \emph{external structure} of input splitting and recombination. When input $\mathbf{z}$ is split into $(\mathbf{z}_1, \mathbf{z}_2)$, the forward transformation is:
\begin{align}
	\mathbf{y}_1 = \mathbf{z}_1, \quad \mathbf{y}_2 = \mathbf{z}_2 \odot \exp\bigl(s(\mathbf{z}_1)\bigr) + t(\mathbf{z}_1),
	\label{eq:acl_fwd}
\end{align}
where $s, t: \mathbb{R}^{d/2} \to \mathbb{R}^{d/2}$ are the scale and shift subnets. Since $\mathbf{y}_1 = \mathbf{z}_1$, the inverse is derived without $s^{-1}$ or $t^{-1}$:
\begin{align}
	\mathbf{z}_1 = \mathbf{y}_1, \quad
	\mathbf{z}_2 = (\mathbf{y}_2 - t(\mathbf{y}_1)) \oslash \exp\bigl(s(\mathbf{y}_1)\bigr).
	\label{eq:acl_inv}
\end{align}
The lower-triangular Jacobian structure yields $\log|\det \mathbf{J}_f(\mathbf{z})| = \sum_{i} s_i(\mathbf{z}_1)$, which depends solely on the outputs of the scale function $s$. This structural independence, ensuring both invertibility and exact likelihood computation regardless of the internal architecture, is formalized as follows:

\begin{proposition}[Subnet-Independent Validity~{\cite{dinh2016realnvp}}]
	\label{prop:subnet_independence}
	For the affine coupling transformation $f: \mathbb{R}^d \to \mathbb{R}^d$ defined by Eq.~\eqref{eq:acl_fwd}, for any differentiable functions $s, t: \mathbb{R}^{d/2} \to \mathbb{R}^{d/2}$:
	\begin{enumerate}
		\item[(i)] $f$ is always invertible regardless of the subnet architecture of $s$ and $t$;
		\item[(ii)] $\log|\det \mathbf{J}_f(\mathbf{z})| = \sum_i s_i(\mathbf{z}_1)$, where the log-det computation formula itself is independent of $t$ and of the internal architecture of $s$.
	\end{enumerate}
\end{proposition}

Since flow validity depends only on subnet \emph{outputs}, the internal weights can be restructured to accommodate continual learning. We exploit this by decomposing each subnet into a shared base and a task-specific low-rank adapter:

\begin{corollary}[Parameter-Level Forgetting Prevention under Correct Routing]
	\label{cor:zero_forgetting}
	Let each subnet be decomposed as $\psi(\mathbf{z}) = \psi_{\mathrm{base}}(\mathbf{z}; \theta_{\mathrm{base}}) + \delta\psi_{\tau}(\mathbf{z}; \Delta\theta_{\tau})$ for $\psi \in \{s, t\}$, where $\theta_{\mathrm{base}}$ is frozen after Task 0, and each task-specific adapter $\Delta\theta_{\tau}$ is stored independently. When adapter $\Delta\theta_{\tau'}$ is loaded at inference, the flow output is identical to the state at the completion of task $\tau'$ training.
\end{corollary}

This decomposition eliminates parameter-induced forgetting by isolating task-specific updates.

\begin{table}[t]
	\centering
	\caption{Cross-architecture comparison under the same frozen base + LoRA strategy.}
	\label{tab:architecture_comparison}
	\begin{tabular}{llccc}
		\toprule
		Architecture           & Decomp. Level           & I-AUC (\%)     & P-AP (\%)      & FM (\%p)      \\
		\midrule
		\textbf{NF (DeCoFlow)} & \textbf{Coupling-level} & \textbf{98.47} & \textbf{58.57} & \textbf{0.00} \\
		NF (Feature-level)     & Feature-level           & 50.00          & 4.15           & 0.00          \\
		Autoencoder            & Decoder LoRA            & 68.15          & 20.52          & 0.00          \\
		VAE                    & Decoder LoRA            & 65.25          & 18.14          & 0.04          \\
		Teacher-Student        & Student LoRA            & 62.03          & 15.34          & 18.98         \\
		\bottomrule
	\end{tabular}
\end{table}

To verify this property is unique to coupling-based NFs, we apply the same frozen+LoRA strategy to other anomaly detection architectures (\cref{tab:architecture_comparison}). AE and VAE yield suboptimal 65--68\% I-AUC because freezing disrupts their encoder-decoder correspondence, while the Teacher-Student architecture suffers from an 18.98\% forgetting rate due to shared encoder updates. Even within NFs, feature-level adaptation causes manifold collapse (I-AUC 50.00\%), as the frozen density mapping cannot accommodate the shifted distribution. Only coupling-level decomposition achieves FM=0.00\% with competitive detection, confirming that subnet independence makes coupling NFs inherently suited for density-based continual learning.

However, a permanently frozen base restricts representational flexibility, biasing the model toward initial task statistics. To overcome this rigidity, the following section introduces compensatory mechanisms within the DeCoFlow framework.


\section{Proposed Method: DeCoFlow}
\label{sec:method}

\subsection{Problem Formulation}
\label{sec:problem_formulation}

Continual Anomaly Detection addresses $T$ tasks, $\mathcal{D} = \{\mathcal{D}_0, \ldots, \mathcal{D}_{T-1}\}$ arriving sequentially. Each task $\mathcal{D}_{\tau}$ contains only normal samples and prior data $\mathcal{D}_{0:\tau-1}$ is inaccessible. We follow the Class-Incremental Learning (CIL) scenario, where Task IDs are unavailable at inference. The NF parameters are denoted as $\theta = \{\theta_{\mathrm{base}}, \{\Delta\theta_{\tau}, \phi_{\tau}\}_{\tau=0}^{T-1}\}$, with $\Delta\theta_{\tau}$ as task-specific LoRA parameters and $\phi_{\tau} = \{\gamma_{\tau}, \beta_{\tau}, \theta_{\mathrm{ACL},\tau}\}$ as auxiliary adaptation parameters (TSA and ACL). For each task $\tau \geq 1$, $\theta_{\mathrm{base}}$ is frozen and only $\Delta\theta_{\tau}, \phi_{\tau}$ are optimized.

\subsection{Framework Overview}
\label{sec:framework_overview}

\begin{figure*}[t]
	\centering
	\includegraphics[width=\textwidth]{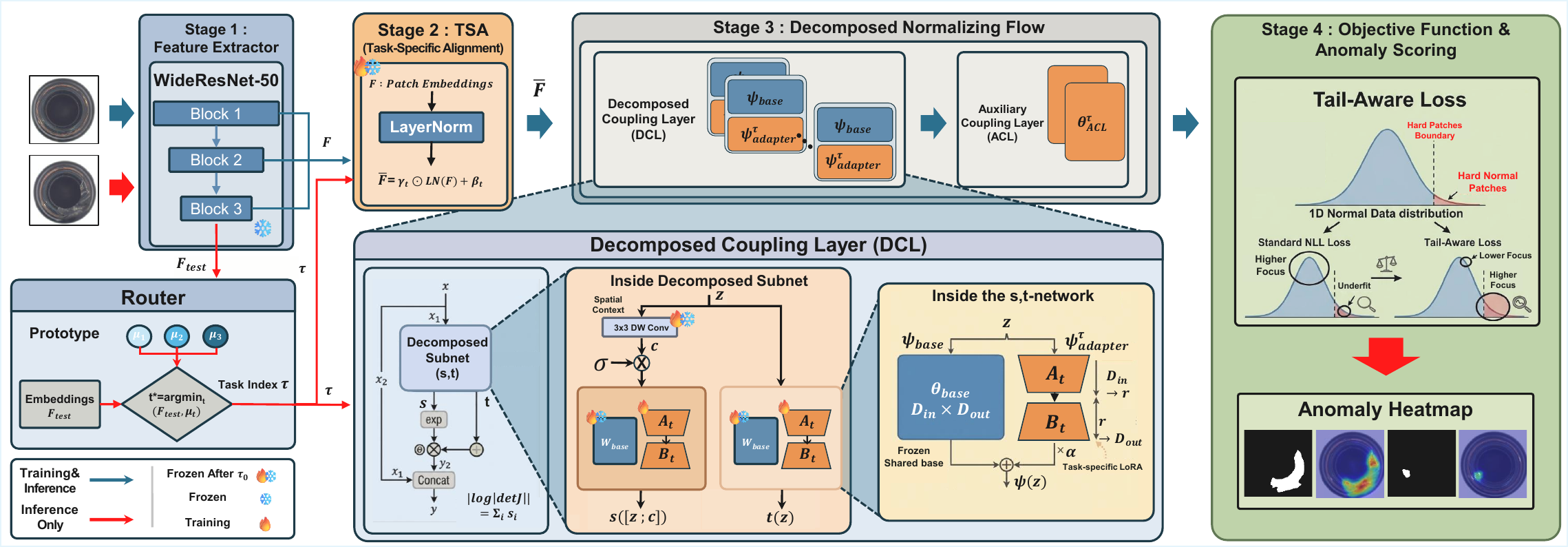}
	\caption{Overall pipeline of DeCoFlow. Top: The sequential flow of multi-scale features through the frozen backbone, TSA calibration, and the Normalizing Flow stages. Middle: Detailed structural views of (A)~the affine coupling process, (B)~the scale subnet architecture with spatial context, and (C)~the parallel $W_{\text{base}} + \Delta W$ decomposition in DCL. Bottom: The training timeline where the base is frozen after Task 0 to enable sequential adapter-only updates for subsequent tasks.}
	\label{fig:main}
\end{figure*}

DeCoFlow (\cref{fig:main}) achieves robust continual anomaly detection by coupling structural isolation with compensatory modules. At its core, the Decomposed Coupling Layer (DCL) partitions subnets into a frozen base and task-specific adapters to eliminate parameter interference. To mitigate frozen-base rigidity, we integrate three components: Task-Specific Alignment (TSA) for input calibration, Auxiliary Coupling Layers (ACL) for residual density correction, and Tail-Aware Loss (TAL) for enhanced learning on challenging samples. During inference, a prototype-based routing mechanism identifies the appropriate adapter via Mahalanobis distance, enabling task-agnostic anomaly detection in a single forward pass.

\subsection{Feature Extraction and Preprocessing}
\label{sec:feature_extraction}

Multi-scale features from a pre-trained backbone are aggregated into unified patch representations $F \in \mathbb{R}^{|\mathcal{B}| \times H \times W \times D}$~\cite{roth2022patchcore}. Unlike previous methods employing separate NFs per scale~\cite{fastflow, gudovskiy2022cflow}, DeCoFlow processes these integrated features through a single NF, minimizing parameter overhead to one LoRA adapter per task. The backbone remains frozen, and we incorporate 2D sinusoidal positional encodings to preserve essential spatial context within the patch grid.

\textbf{Task-Specific Alignment (TSA)} The NF base weights, frozen after Task 0 (\cref{sec:theoretical_foundation}), remain biased toward initial task statistics, potentially making subsequent inputs appear out-of-distribution. TSA mitigates this shift by standardizing patch features via LayerNorm followed by a task-specific affine transformation $(\gamma_{\tau}, \beta_{\tau})$:
\begin{equation}
	f_{\text{LN}} = \frac{F - \mathbb{E}[F]}{\sqrt{\text{Var}[F] + \epsilon}}, \quad \hat{F}_{\tau} = \gamma_{\tau} \odot f_{\text{LN}} + \beta_{\tau}.
	\label{eq:tsa_affine}
\end{equation}
The parameters $\gamma_{\tau}$ and $\beta_{\tau}$ are constrained by sigmoid and tanh activations, respectively, to ensure stable and bounded outputs. Notably, TSA serves as an external preprocessing step rather than an inherent flow component. Because the NF treats $\hat{F}_{\tau}$ as a static input, LayerNorm statistics ($\mathbb{E}[F], \text{Var}[F]$) do not impact the log-determinant, thus preserving the coupling layers' original Jacobian structure.

\subsection{Decomposed Coupling Layer}
\label{sec:dcl}

DCL serves as the core building block of DeCoFlow. By partitioning each coupling subnet into a frozen shared base and a task-specific adapter, it composes the subnets as the sum of these components, following the principles established in \cref{sec:theoretical_foundation}.

\textbf{Internal Subnet Structure} Each subnet $\psi \in \{s, t\}$ is decomposed into a frozen base and a task-specific adapter via an additive mapping:
\begin{equation}
	\psi(\mathbf{z}) = \underbrace{\psi_{\mathrm{base}}(\mathbf{z}; \theta_{\mathrm{base}})}_{\text{shared across tasks}} + \underbrace{\delta\psi_{\tau}(\mathbf{z}; \Delta\theta_{\tau})}_{\text{task-specific}},
	\label{eq:subnet_structure}
\end{equation}
where the adapter $\delta\psi_{\tau}$ is implemented via LoRA~\cite{lora} with zero-initialized $\mathbf{B}_{\tau}$. This ensures that each task initially inherits the frozen base distribution. Per \cref{prop:subnet_independence}, this decomposition preserves invertibility and Jacobian computation, as they remain invariant to the internal architecture of $\psi$.

\textbf{Asymmetric Design with Context Injection} While standard NFs typically employ symmetric architectures for both subnets, we adopt an asymmetric configuration to better align with their distinct roles in density-based anomaly detection. Specifically, the scale function ($s$) is responsible for localized defect capture through volume change, whereas the shift function ($t$) handles global distribution translation. This functional distinction motivates the following differentiated structures:
\begin{equation}
s([\mathbf{z}; \mathbf{c}]) = s_{\mathrm{base}}([\mathbf{z}; \mathbf{c}]) + \delta s_{\tau}([\mathbf{z}; \mathbf{c}]). \label{eq:dcl_scale}
\end{equation}
Conversely, to ensure stable translation and noise robustness, $t$ operates strictly on $\mathbf{z}$ without external context:
\begin{equation}
t(\mathbf{z}) = t_{\mathrm{base}}(\mathbf{z}) + \delta t_{\tau}(\mathbf{z}). \label{eq:dcl_shift}
\end{equation}

To aggregate neighborhood information, $\mathbf{c}$ is computed by applying a $3\times3$ depthwise convolution on the spatially-reshaped 2D patch grid, followed by flattening the result back to 1D. The spatial context is further modulated by a learned gate $\eta = \eta_{\max}\,\sigma(\theta_c)$, which is frozen after Task 0 to ensure that the spatial augmentation remains consistent across all subsequent tasks.

\textbf{Continual Learning Protocol} In Task 0, $\theta_{\mathrm{base}}$ and $\Delta\theta_0$ are jointly trained, after which the base is permanently frozen. For $\tau \geq 1$, only LoRA $\Delta\theta_{\tau}$ and TSA $\phi_{\tau}$ are optimized, with $\mathbf{B}_{\tau}$ zero-initialized. This freezing strategy is justified by a high inter-task gradient cosine similarity of 0.78 (\cref{tab:mechanism}c), indicating that the learned base features are sufficiently task-invariant.

\subsection{Auxiliary Coupling Layers (ACL)}
\label{sec:acl}
Due to the structural constraint of affine coupling---where only half of the dimensions are transformed per step---complete cross-dimensional decorrelation is not inherently guaranteed. This is further compounded by the low-rank nature of LoRA, which restricts statistical adjustments to a rank-$r$ subspace and may prevent full-rank correction across all dimensions. Consequently, residual correlations can accumulate through the DCL stack, a phenomenon empirically analyzed in \cref{sec:mechanism_analysis} (\cref{fig:blockwise_enhanced}b). To bridge this statistical gap, we introduce Auxiliary Coupling Layers (ACL) to provide additional residual density correction.

Auxiliary Coupling Layers (ACL) are a small number of affine coupling layers appended after the DCLs to compensate for these residual statistical discrepancies:
\begin{equation}
	\mathbf{z}_{\mathrm{out}} = g_{\mathrm{ACL},\tau}(\mathbf{z}_{\mathrm{DCL}};\,
	\theta_{\mathrm{ACL},\tau}).
	\label{eq:acb}
\end{equation}
To bridge the remaining statistical gap, each task is assigned independent ACL parameters $\theta_{\mathrm{ACL},\tau}$, as residual correlations are inherently task-specific. Each ACL subnet is implemented as a two-layer MLP with a hidden dimension $d_{\mathrm{hidden}} = \lfloor D/2 \rfloor$. Crucially, the output layer is zero-initialized so that the module initially acts as an identity function. This allows ACL to introduce progressive statistical corrections without perturbing the density transformation already established by the DCL stack.

While adding ACL entails parameter overhead, it remains far more efficient than allocating independent networks for each task. Crucially, the statistical normalization in ACL is structurally complementary to DCL's nonlinear mapping, such that its role cannot be substituted by merely increasing the number of DCL blocks. Detailed analyses of this efficiency and modular synergy are provided in \cref{sec:mechanism_analysis}.

\subsection{Training Objective: Tail-Aware Loss (TAL)}

DeCoFlow is trained by minimizing the Negative Log-Likelihood (NLL) derived from the change-of-variables formula:
\begin{equation}
	\mathcal{L}_{\text{NLL}} = -\log p_Z(\mathbf{z}) - \log|\det \mathbf{J}_f(\mathbf{x})| = \tfrac{1}{2}\|\mathbf{z}\|^2 - \log|\det \mathbf{J}_f(\mathbf{x})|,
	\label{eq:nll}
\end{equation}
where $\mathbf{z} = f(\mathbf{x})$ is the latent output and $p_Z = \mathcal{N}(\mathbf{0}, \mathbf{I})$ is the standard Gaussian prior (constant omitted). However, standard NLL optimization often prioritizes dominant high-likelihood modes, causing limited adapter capacity to be over-allocated to the distribution center. This results in underfitting low-density tail regions, which are critical for distinguishing normal samples from anomalies.

To mitigate this capacity imbalance, we introduce Tail-Aware Loss (TAL), drawing inspiration from hard-example mining techniques such as OHEM~\cite{shrivastava2016ohem} and Focal Loss~\cite{lin2017focal}. TAL reweights high-loss patches to redirect the constrained representational capacity of LoRA subnets toward under-represented tail distributions:
\begin{equation}
    \mathcal{L}_{\text{TAL}} = (1 - \omega)\, \mathbb{E}_{x \in \mathcal{B}} [\mathcal{L}_{\text{NLL}}(x)] + \omega\, \mathbb{E}_{x \in \mathcal{K}} [\mathcal{L}_{\text{NLL}}(x)],
    \label{eq:tal}
\end{equation}
where $\omega$ is the tail weight ratio, $\mathcal{B}$ is the full batch, and $\mathcal{K}$ denotes the set of top-$k$ highest NLL-loss patches with $k = \lfloor |\mathcal{B}|HW \cdot r_{\text{tail}} \rfloor$. We set $\omega=0.85$ for MVTec-AD, $\omega=0.8$ for VisA, and $r_{\text{tail}}=0.02$. These parameters remain fixed across all tasks as the tail ratio of normal distributions is empirically stable, while dynamic adjustment remains a subject for future work.

Since TAL amplifies gradients on tail patches, scale function outputs can lead to uncontrolled volume expansion. To ensure numerical stability and stable density estimation on the frozen base, we introduce an $\ell_2$ regularization term on the log-Jacobian determinant. The final objective is defined as:
\begin{equation}
    \mathcal{L}_{\text{total}} = \mathcal{L}_{\text{TAL}} + \lambda_{\text{reg}} \left\| \log|\det \mathbf{J}_f(\mathbf{z})| \right\|^2_2,
    \label{eq:total_loss}
\end{equation}
where $\lambda_{\text{reg}} = 10^{-4}$ serves to constrain the Jacobian magnitude.

\subsection{Task Routing and Inference}
\label{sec:routing}

\textbf{Prototype-Based Task Routing} To identify the appropriate task-specific adapter without auxiliary networks, we employ prototype-based matching. For each task $\tau$, a prototype is defined by the mean $\mu_{\tau}$ and covariance $\Sigma_{\tau}$ of normal feature statistics computed from the training set. At inference, the routing mechanism selects the optimal task index $\tau^*$ by minimizing the Mahalanobis distance between the test feature $f_{\text{test}}$ and the stored prototypes:
\begin{equation}
	\tau^* = \arg\min_{\tau} \sqrt{(f_{\text{test}} - \mu_{\tau})^T \Sigma_{\tau}^{-1} (f_{\text{test}} - \mu_{\tau})}.
	\label{eq:mahalanobis}
\end{equation}

Specifically, we use only the final scale of the multi-scale features as the routing query $f_{\text{test}}$ to capture the global task context. Although the subsequent NF process utilizes multi-scale features to preserve dense structural details, routing relies solely on this high-level semantic representation. This strategy ensures accurate task identification within a single forward pass, effectively eliminating the requirement for an external classifier or additional backbone computations.

Since the score is computed through the routed adapter, correct routing is required to realize the zero-forgetting property: misrouting does degrade performance, as forcing 5\% misrouting on MVTec-AD lowers P-AP from 56.93\% to 30.20\%. In practice, however, every class retains a sufficient routing margin---even the worst case (VisA \emph{pcb2}) keeps its second-nearest prototype $2.58\times$ farther than the nearest---so Mahalanobis routing reaches 100\% accuracy on both datasets.

\textbf{Anomaly Scoring} Patch-level anomaly scores are defined as NLL $a_{h,w} = -\log p(z_{h,w}) - \log|\det \mathbf{J}_{h,w}|$. The image-level score aggregates the top-$K$ patches such that $a_{\text{img}} = \frac{1}{K}\sum_{(h,w) \in \mathcal{T}_K} a_{h,w}$, where $\mathcal{T}_K$ denotes the set of $K=3$ patches with the highest anomaly scores. This top-$K$ strategy is particularly suited for manufacturing data where anomalies are locally concentrated, as it prevents the dilution of anomalous signals by surrounding normal regions.


\section{Experiments}
\label{sec:experiments}

\subsection{Experimental Setup}

\textbf{Datasets and Protocol} We evaluate DeCoFlow on MVTec-AD~\cite{bergmann2019mvtec} and VisA~\cite{zou2022spot} benchmarks. All experiments follow the target 1$\times$1 CIL scenario: classes are learned sequentially under zero-replay constraints and task-agnostic inference. To ensure consistent comparison, we apply an identical alphabetical learning order for all methods across both benchmarks.

\textbf{Baselines and Metrics} We compare against joint-training (PatchCore~\cite{roth2022patchcore}, CADIC~\cite{cadic}), fine-tuning (PatchCore, CFA~\cite{cfa2022}, SimpleNet~\cite{simplenet2023}, RD4AD~\cite{rd4ad2022}), and diverse CL baselines (EWC~\cite{ewc}, LwF~\cite{li2017lwf}, Replay, IUF~\cite{iuf2024}, ReplayCAD~\cite{replaycad}, CDAD~\cite{cdad2025}, DNE~\cite{dne}, UCAD~\cite{ucad}, CADIC~\cite{cadic}). Primary evaluation metrics are I-AUC and P-AP, with FM ($\downarrow$) and routing accuracy as auxiliary measures.

\textbf{Implementation Details} For MVTec-AD, features are extracted from a WideResNet-50-2 backbone. The NF is configured with 6 DCL and 2 ACL blocks at LoRA rank 16, then trained via AdamP (lr=$3 \times 10^{-4}$) for 60 epochs with a batch size of 16. For VisA, we utilize ViT-B/16 features from blocks [1,2,3,5] with 10 DCL and 6 ACL blocks (lr=$2 \times 10^{-4}$) for 100 epochs. Both benchmarks train the base on Task 0, freeze it, and subsequently update only the adapters. Results denote the mean and standard deviation over three seeds at 224$\times$224 resolution; full-stream training takes about 3.0 GPU-hours on MVTec-AD and 4.1--4.4 GPU-hours on VisA.

\subsection{Main Results}
\label{sec:main_results}

\begin{table*}[t]
	\centering
	\caption{Performance comparison on MVTec-AD and VisA in the 1$\times$1 CIL setting. Entries denote average performance (Avg) and forgetting measure (FM) after the final task. \textbf{Bold}/\underline{underlined} values indicate best/second-best results. \emph{Joint} denotes reference baselines trained on all tasks simultaneously. $^\dagger$ signifies parameter-level zero forgetting under correct routing; $^\ddagger$ results are from \cite{cadic}. All values are in \%.}
	\label{tab:main_results_compact}
	\resizebox{\textwidth}{!}{%
	\begin{tabular}{l cccc cccc}
			\toprule
			\multirow{3}{*}{Method}                       & \multicolumn{4}{c}{MVTec-AD (15-class)} & \multicolumn{4}{c}{VisA (12-class)}                                                                                                                                                     \\
			\cmidrule(lr){2-5} \cmidrule(lr){6-9}
			                                              & \multicolumn{2}{c}{I-AUC}               & \multicolumn{2}{c}{P-AP}            & \multicolumn{2}{c}{I-AUC} & \multicolumn{2}{c}{P-AP}                                                                                              \\
			\cmidrule(lr){2-3} \cmidrule(lr){4-5} \cmidrule(lr){6-7} \cmidrule(lr){8-9}
			                                              & Avg ($\uparrow$)                        & FM ($\downarrow$)                   & Avg ($\uparrow$)          & FM ($\downarrow$)        & Avg ($\uparrow$)      & FM ($\downarrow$)      & Avg ($\uparrow$) & FM ($\downarrow$)      \\
			\midrule
			\emph{Joint\_PatchCore}                       & \emph{97.80}                            & \emph{--}                           & \emph{59.40}              & \emph{--}                & \emph{91.60}          & \emph{--}              & \emph{44.00}     & \emph{--}              \\
			\emph{Joint\_CADIC}                           & \emph{97.80}                            & \emph{--}                           & \emph{59.10}              & \emph{--}                & \emph{91.00}          & \emph{--}              & \emph{43.90}     & \emph{--}              \\
			\midrule
			FT\_PatchCore$^\ddagger$                      & 60.20                                   & 38.30                               & 19.00                     & 37.10                    & 58.90                 & 36.10                  & 9.00             & 31.10                   \\
			FT\_CFA$^\ddagger$~\cite{cfa2022}             & 62.30                                   & 36.10                               & 17.70                     & 8.30                     & 59.30                 & 32.70                  & 8.70             & 18.40                   \\
			FT\_SimpleNet$^\ddagger$~\cite{simplenet2023} & 70.80                                   & 21.10                               & 6.00                      & 6.90                     & 61.60                 & 28.30                  & 1.40             & 1.60                    \\
			FT\_RD4AD$^\ddagger$~\cite{rd4ad2022}         & 59.60                                   & 39.30                               & 14.30                     & 42.50                    & 52.50                 & 42.30                  & 6.90             & 20.10                   \\
			\midrule
			IUF~\cite{iuf2024}                            & 76.20                                   & 6.70                                & 17.10                     & 5.90                     & 68.10                 & 8.50                   & 3.40             & \underline{0.30}        \\
			ReplayCAD~\cite{replaycad}                    & 94.80                                   & 4.50                                & 53.70                     & 5.50                     & \underline{90.30}     & 5.50                   & \underline{41.50}& 5.00                    \\
			CDAD~\cite{cdad2025}                          & 74.30                                   & 20.10                               & 28.00                     & 26.10                    & 62.80                 & 22.20                  & 8.30             & 21.70                   \\
			DNE~\cite{dne}                                & 87.00                                   & 11.60                               & --                        & --                       & 61.00                 & 17.90                   & --               & --                     \\
			UCAD~\cite{ucad}                              & 93.00                                   & \underline{1.00}                    & 45.60                     & \underline{1.30}         & 87.40                 & \underline{3.90}       & 30.00            & 1.50                    \\
			CADIC~\cite{cadic}                            & \underline{97.20}                       & 1.10                                & \textbf{58.40}            & 1.50                     & 89.10                 & 4.30                   & \textbf{43.80}   & 1.40                    \\
			\midrule
			DeCoFlow (Ours)                               & \textbf{98.40}$\pm$0.00                 & \textbf{0.00}$^\dagger$             & \underline{58.20}$\pm$0.10& \textbf{0.00}$^\dagger$  & \textbf{93.00}$\pm$0.10& \textbf{0.00}$^\dagger$& 37.00$\pm$0.30   & \textbf{0.00}$^\dagger$ \\
			\bottomrule
	\end{tabular}}
\end{table*}

\textbf{MVTec-AD} DeCoFlow achieves 98.40\% I-AUC, surpassing the joint reference baselines in image-level detection, with competitive 58.20\% P-AP. It maintains parameter-level zero forgetting (FM=0.00\%) by structurally bypassing inter-task parameter interference (\cref{cor:zero_forgetting}): unlike replay- or regularization-based methods, DeCoFlow secures stability through structural parameter isolation rather than constrained updates. On MVTec-AD, Mahalanobis routing also reaches 100\% accuracy. Performance is stable across class orders, with alphabetical, reverse, and random streams yielding I-AUC 98.47/98.43/98.40 and P-AP 58.57/59.20/58.23 at FM=0.00\%.

\begin{figure}[t]
	\centering
	\includegraphics[width=\columnwidth]{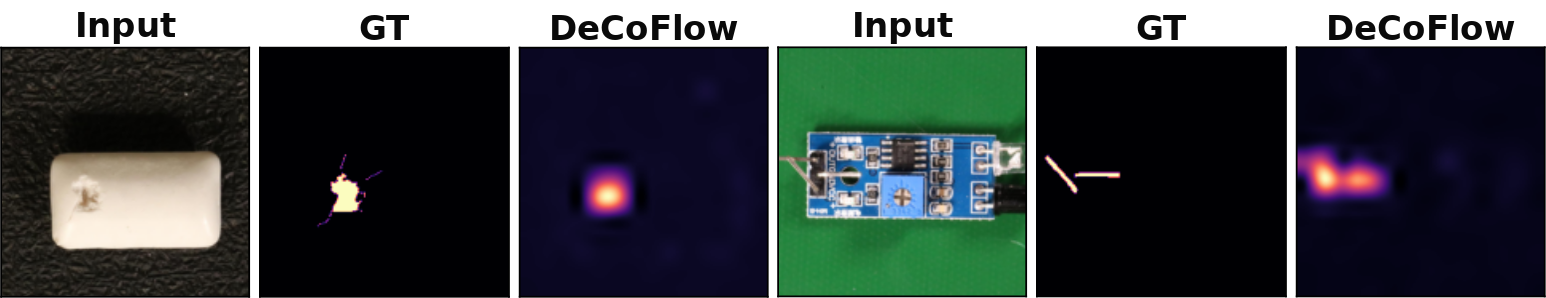}
	\caption{VisA fine-defect localization failures (Input, GT, DeCoFlow). Defect regions are detected, but thin or low-contrast boundaries are blurred---consistent with the fine-defect P-AP gap rather than a density-modeling failure.}
	\label{fig:visa_failure}
\end{figure}

\textbf{VisA} On VisA, DeCoFlow attains 93.00\% I-AUC with zero forgetting, outperforming ReplayCAD by +2.70\%p and CADIC by +3.90\%p in detection. Pixel-level localization is weaker, with P-AP at 37.00\% against CADIC's 43.80\%, and the gap concentrates on fine-defect classes such as \emph{chewinggum} ($\Delta$P-AP $-51.3$) and \emph{pcb3} ($-28.8$), where thin, low-contrast boundaries are oversmoothed (\cref{fig:visa_failure}). This behavior traces to feature granularity rather than the density model: the ViT-B/16 backbone raises I-AUC over WideResNet-50-2 (93.00\% vs.\ 88.60\%) but yields a coarse $14\times14$ grid, and raising token resolution lifts P-AP from 35.85\% to 39.52\%, while stronger backbones reach I-AUC/P-AP 95.77/44.01 (EVA02-B14) and P-AP 38.05 (ViT-B/8). Detection thus remains robust, with localization the primary VisA limitation.

\begin{figure*}[t]
	\centering
	\includegraphics[width=\textwidth]{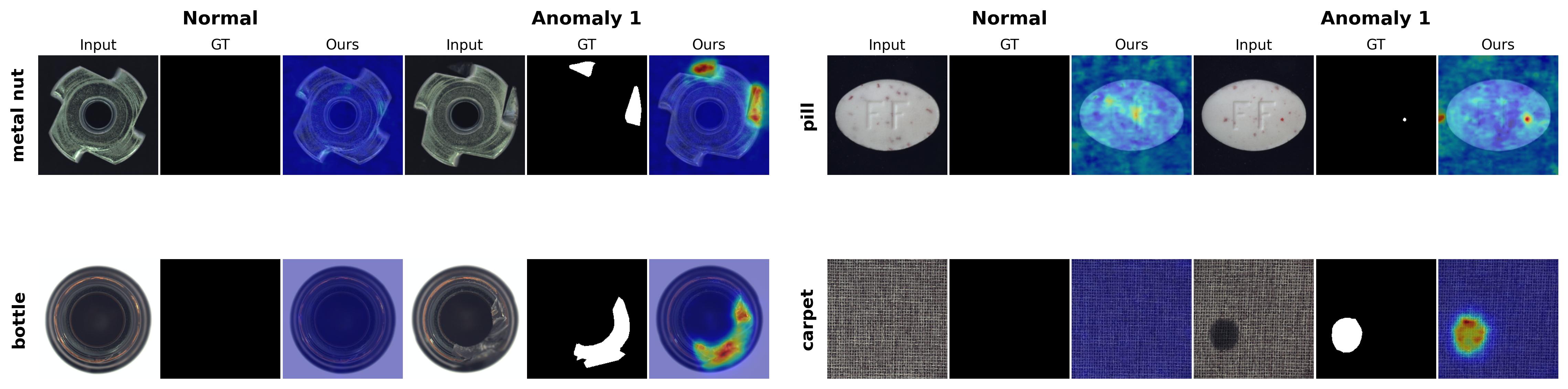}
	\caption{Qualitative anomaly maps for four representative classes (one normal and two anomalous samples per class). DeCoFlow assigns consistently low scores to normal regions and localized high responses to defect regions.}
	\label{fig:anomaly_maps}
\end{figure*}

\textbf{Qualitative Evidence} To complement quantitative metrics, \cref{fig:anomaly_maps} provides qualitative anomaly maps. Across representative classes, DeCoFlow assigns low scores to normal regions and concentrates high responses on defect areas, consistent with the image-level gains in \cref{tab:main_results_compact}.

\begin{figure}[t]
	\centering
	\includegraphics[width=\columnwidth]{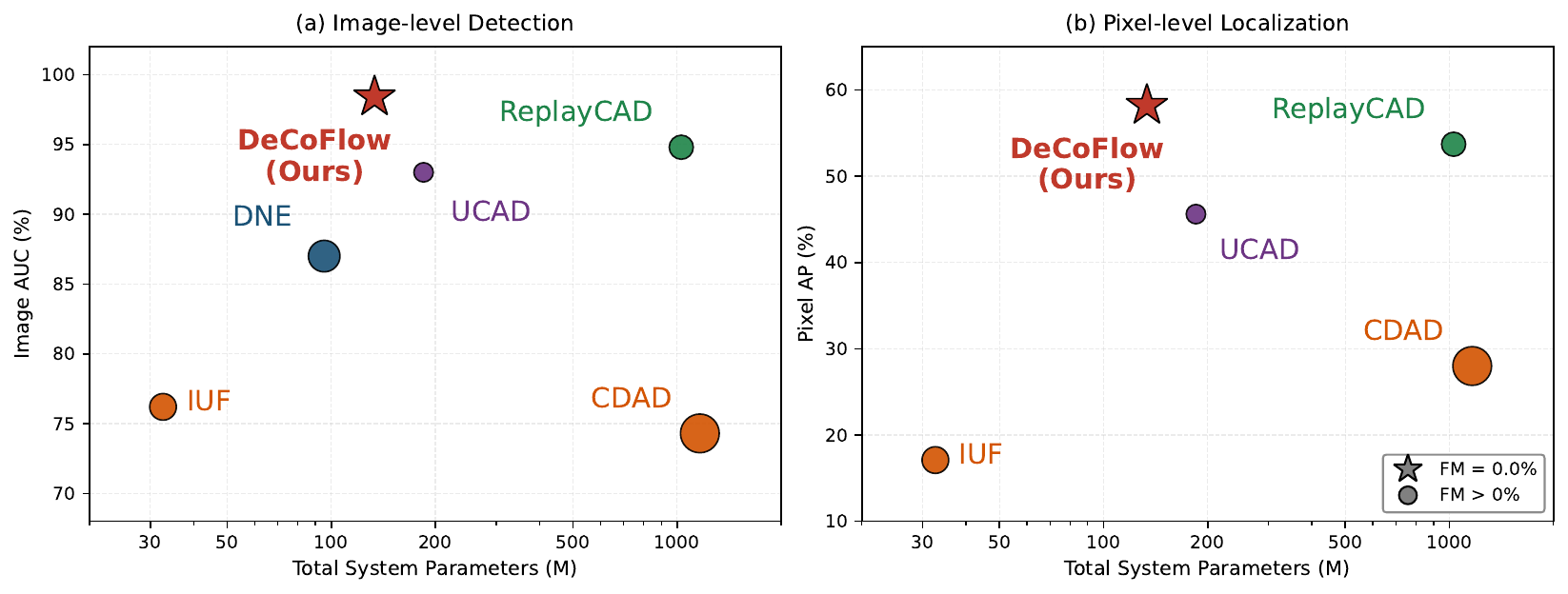}
	\caption{Performance, parameter cost, and forgetting comparison on MVTec-AD 15-class.}
	\label{fig:scatter_efficiency}
\end{figure}

\textbf{Structural Cost Analysis} \Cref{fig:scatter_efficiency} positions DeCoFlow against baselines in the accuracy--parameter--forgetting trade-off. \Cref{tab:design_space} confirms that updating the shared base (up to 25.0$\times$ cost) or employing high-rank adapters (4.7--7.5$\times$ overhead) yields no significant gains. DeCoFlow adopts rank=16, aligned with the SVD effective rank of 17.23 (\cref{tab:mechanism}a), requiring only 2.27M parameters per task (2.5\%). End-to-end inference takes 90.2\,ms/image (11.1 FPS) with 989\,MB peak memory on the 15-task MVTec-AD stream, faster than CADIC's 10k-coreset setting (6.6 FPS) while storing parameters rather than coresets.

\subsection{Ablation Study}
\label{sec:component_analysis}

\begin{figure}[t]
\begin{minipage}[t]{0.58\linewidth}
	\centering
	\captionof{table}{Design space comparison (MVTec-AD): varying adapter structure and configuration on the same NF backbone. Rel.\ denotes parameter ratio relative to DeCoFlow.\newline}
	\label{tab:design_space}
	\resizebox{\linewidth}{!}{%
		\begin{tabular}{@{}llcccc@{}}
			\toprule
			Design                        & Adapter                  & I-AUC          & P-AP           & Params        & Rel.                 \\
			                              &                          & (\%)           & (\%)           & (M)           &                      \\
			\midrule
			Full ACL (8b, $h{=}0.5$)      & Full/task                & 98.74          & 58.27          & 7.10          & 3.1$\times$          \\
			Full ACL (8b, $h{=}2.0$)      & Full/task                & 98.83          & 57.81          & 56.68         & 25.0$\times$         \\
			DCL + Linear                  & Full-rank                & 98.48          & 57.78          & 10.64         & 4.7$\times$          \\
			DCL + LoRA ($r{=}512$)        & LoRA                     & 98.46          & 58.56          & 17.12         & 7.5$\times$          \\
			DCL-only 6b (w/o ACL)         & LoRA ($r{=}16$)          & 88.30          & 45.76          & 1.49          & 0.66$\times$         \\
			DCL-only 8b (w/o ACL)         & LoRA ($r{=}16$)          & 89.46          & 47.38          & 2.12          & 0.93$\times$         \\
			\midrule
			\textbf{DeCoFlow (6DCL+2ACL)} & \textbf{LoRA ($r{=}16$)} & \textbf{98.49} & \textbf{58.57} & \textbf{2.27} & \textbf{1.0$\times$} \\
			\bottomrule
		\end{tabular}}
\end{minipage}%
\hfill
\begin{minipage}[t]{0.40\linewidth}
	\centering
	\captionof{table}{Component ablation results (MVTec-AD). $^\ddag$w/o LoRA entails 100\% parameter increase per task.\newline}
	\label{tab:ablation_component}
	\resizebox{\linewidth}{!}{%
		\begin{tabular}{@{}lccc@{}}
			\toprule
			Configuration    & I-AUC (\%) & P-AP (\%) & $\Delta$ I-AUC \\
			\midrule
			Full (DeCoFlow)  & 98.47      & 58.57     & --             \\
			w/o TSA + TAL    & 94.18      & 48.56     & $-$4.29        \\
			w/o ACL          & 88.30      & 45.76     & $-$10.17       \\
			w/o TAL          & 95.30      & 49.67     & $-$3.17        \\
			w/o TSA          & 97.79      & 55.00     & $-$0.68        \\
			w/o LoRA$^\ddag$ & 98.43      & 58.90     & $-$0.04        \\
			\bottomrule
		\end{tabular}}
\end{minipage}
\end{figure}

\begin{table}[t]
	\centering
	\caption{Factorial analysis: I-AUC change (\%p) per component under frozen vs.\ trainable base regimes. $\Delta$ is measured from each regime's uncompensated baseline (Frozen: 84.12\%, Trainable: 68.13\%). $2 \times 4$ design to characterize regime-dependent behavior.}
	\label{tab:interaction_effect}
	\begin{tabular}{lccc}
		\toprule
		Component & Frozen ($\Delta$\%p) & Trainable ($\Delta$\%p) & Regime Dep.            \\
		\midrule
		TAL       & +13.04               & $-$1.22                 & Frozen-specific        \\
		ACL       & +11.52               & +6.93                   & 1.66$\times$ amplified \\
		TSA       & $-$1.19              & $-$16.34                & Suppressed             \\
		\bottomrule
	\end{tabular}
\end{table}

\textbf{Components Analysis} Coupling decomposition alone (without TSA and TAL) achieves 94.18\%, establishing DCL as the primary architectural component (\cref{tab:ablation_component}). The performance gap between this baseline and the full model reflects frozen-base rigidity, which each auxiliary module mitigates through a specialized function. Specifically, removing ACL incurs the largest degradation ($-$10.17\%p), indicating that DCL requires explicit normalization to resolve residual statistical mismatches (\cref{fig:blockwise_enhanced}). Similarly, the absence of TAL causes underfitting in low-density tail regions ($-$3.17\%p), while removing TSA compromises feature alignment for subsequent tasks ($-$0.68\%p). Finally, excluding LoRA results in a negligible accuracy impact ($-$0.04\%p) but doubles the parameter count per task, confirming its role as a critical efficiency mechanism.

\textbf{Regime-Specific Interaction Analysis} Factorial analysis in \cref{tab:interaction_effect} reveals that our compensators are specifically tailored for the frozen NF base regime rather than being universally beneficial modules. TAL provides substantial gains (+13.04\%p) when the NF base weights are frozen but becomes redundant or even deleterious ($-$1.22\%p) when the base is trainable, validating its specialized role as a capacity reallocation mechanism. While ACL improves performance in both settings, its contribution is 1.66$\times$ greater under a frozen NF base. TAL is also locally stable: TailTopK/TailW sweeps vary I-AUC only within 97.83--97.92\% and P-AP within 56.03--56.18\%. Finally, TSA contributes primarily through synergistic interaction with other modules rather than through standalone gains.

\subsection{Mechanism Analysis}
\label{sec:mechanism_analysis}

\begin{table}[t]
	\centering
	\caption{Mechanism analysis. Part~A: Low-rank sufficiency (24 $\Delta W$ matrices, 15-task average). Part~B: TAL gradient redistribution (4 tasks $\times$ 30 batches average). Part~C: Task-agnostic base validation (5 different Task~0 initializations).}
	\label{tab:mechanism}
	\begin{minipage}[t]{0.58\linewidth}
		\resizebox{\linewidth}{!}{%
			\begin{tabular}{@{}lccc@{}}
				\toprule
				\multicolumn{4}{@{}l}{\textbf{Part A: Low-Rank Sufficiency}}                            \\
				\midrule
				Metric                     & Value           & \multicolumn{2}{c}{Perf.}                \\
				\midrule
				Eff. Rank (SVD)            & 17.23$\pm$1.10  & \multicolumn{2}{c}{---}                  \\
				Energy@16                  & 92.82\%         & \multicolumn{2}{c}{98.49 / 58.57}        \\
				Energy@64                  & 99.71\%         & \multicolumn{2}{c}{98.47 / 58.57}        \\
				RAR ($\|\Delta W\|/\|W\|$) & 0.092$\pm$0.027 & \multicolumn{2}{c}{---}                  \\
				\midrule\addlinespace[3pt]
				\multicolumn{4}{@{}l}{\textbf{Part B: TAL Gradient Redistribution}}                     \\
				\midrule
				Loss Config                & Grad (Tail)     & Grad (Non-Tail) & Ratio                  \\
				\midrule
				Mean-only                  & 0.00296         & 0.00212         & 1.38$\times$           \\
				Tail-Aware                 & 0.1104          & 0.000755        & 144.2$\times$          \\
				Amplification              & 37.1$\times$    & 0.36$\times$    & 104.5$\times$          \\
				\bottomrule
			\end{tabular}}
	\end{minipage}%
	\hfill
	\begin{minipage}[t]{0.40\linewidth}
		\resizebox{\linewidth}{!}{%
			\begin{tabular}{@{}lcc@{}}
				\toprule
				\multicolumn{3}{@{}l}{\textbf{Part C: Task-Agnostic Base}}           \\
				\midrule
				Metric    & NF                     & MLP                                \\
				\midrule
				Cos-Sim   & \textbf{0.78}$\pm$0.06 & 0.27$\pm$0.11                      \\
				I-AUC Std & 0.21\%p                & ---                                \\
				Routing   & 100\%                  & ---                                \\
				\bottomrule
			\end{tabular}}
	\end{minipage}
\end{table}

\Cref{tab:mechanism} analyzes the principles behind DeCoFlow's precise task adaptation with few parameters from two perspectives.

\textbf{Low-Rank Sufficiency} SVD analysis of $\Delta W$ after 15-task training (\cref{tab:mechanism}a) reveals a mean effective rank of 17.23. Our configuration of rank 16 directly aligns with this inherent low-dimensional structure, capturing the essential signal required for SOTA performance with minimal overhead. This sufficiency is further supported by the high inter-task gradient similarity (cos-sim 0.78, \cref{tab:mechanism}c), which suggests that task adaptation occurs within a compact subspace. Additionally, the relative adaptation ratio (RAR $= \|\Delta W\|_F / \|W_{\text{base}}\|_F$) averages only 0.092, confirming that the frozen base handles 90.8\% of the overall transformation while the low-rank adapters manage task-specific adjustments. A rank sweep from $r{=}16$ to $128$ changes I-AUC/P-AP by only 0.02/0.09\%p, indicating that performance is governed more by the DCL/ACL boundary and TAL objective than by raw adapter rank.

\textbf{Tail Gradient Amplification} Under the frozen-base regime, we leverage the limited capacity of low-rank adapters through strategic gradient allocation rather than relying on parameter-intensive full-rank updates. Standard NLL optimization produces a near-uniform gradient distribution, with a tail-to-non-tail ratio of only $1.38\times$ (\cref{tab:mechanism}b). Conversely, TAL significantly amplifies this ratio to \textbf{144.2$\times$}, representing a $104.5$-fold increase. This shift effectively concentrates the available parameter budget on critical low-density boundaries where normal and anomalous distributions overlap. This redistribution enables a +3.17\%p I-AUC gain (\cref{tab:ablation_component}) without overhead, showing targeted optimization can outperform capacity scaling in continual learning.

\begin{figure*}[t]
	\centering
	\includegraphics[width=0.95\textwidth]{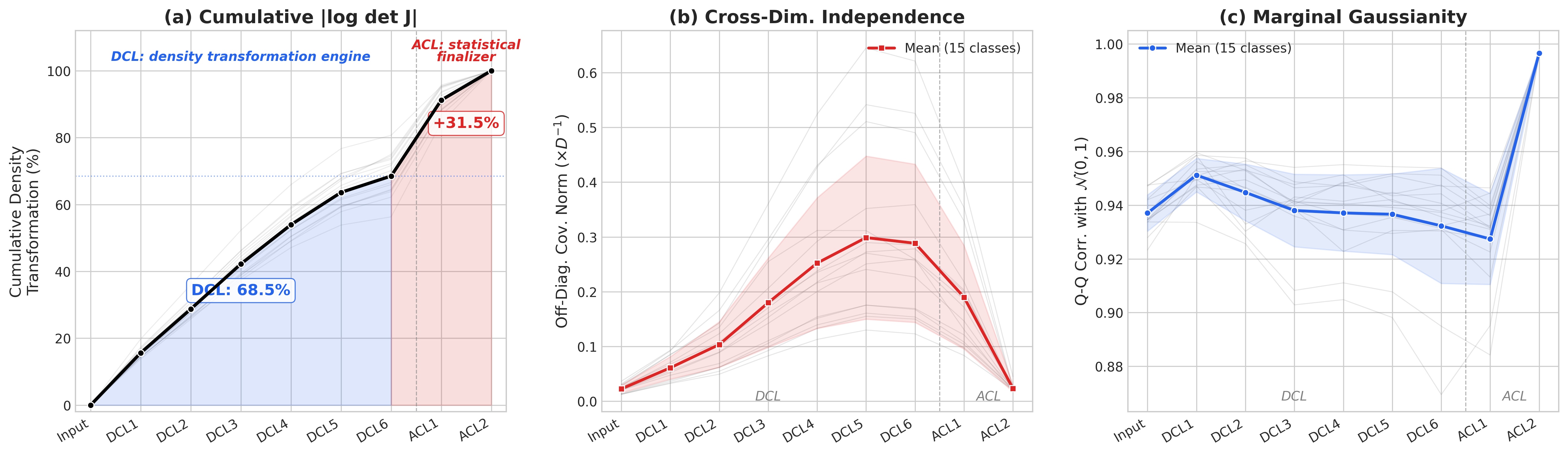}
	\caption{\textbf{Block-wise transformation analysis.} (a)~Cumulative $|\log|\det \mathbf{J}_f(\mathbf{z})||$: DCL 6 blocks contribute 68.5\%, ACL 2 blocks contribute 31.5\%. (b)~Off-Diagonal Covariance: DCL introduces cross-dimensional coupling, which ACL restores to independence. (c)~Q-Q Correlation: stagnates through DCL, then sharply aligns after ACL. Gray lines denote individual classes (15), bold line denotes the mean.}
	\label{fig:blockwise_enhanced}
\end{figure*}

\textbf{Block-Wise Complementarity} \cref{fig:blockwise_enhanced} illustrates how DCL and ACL operate in tandem to achieve precise density modeling. While DCL serves as the primary engine by executing 68.5\% of the total transformation (\cref{fig:blockwise_enhanced}a), it inevitably induces cross-dimensional coupling, as evidenced by a rise in off-diagonal covariance (\cref{fig:blockwise_enhanced}b). ACL subsequently acts as a statistical finalizer, restoring independence and ensuring sharp marginal Gaussian alignment (\cref{fig:blockwise_enhanced}c). This complementarity is not substitutable: replacing ACL with additional DCL blocks (8DCL+0ACL) yields a 9.01\%p performance deficit despite maintaining identical depth. These results confirm that DCL handles the heavy lifting of transformation while ACL provides the necessary statistical regularization for accurate anomaly detection.


\textbf{Task-Agnostic Base Validation} The frozen-base strategy assumes that the initial Task~0 base generalizes across all subsequent tasks (\cref{tab:mechanism}c). To validate this, we independently trained Task~0 on five different classes and measured the pairwise cosine similarity of their resulting gradients. The average similarity reached \textbf{0.78}, indicating highly aligned optimization trajectories across different classes. As a control, a standard MLP yielded a similarity of only 0.27 ($p < 0.003$, Welch's $t$-test), confirming that this alignment is an intrinsic property of the affine coupling architecture when paired with a shared Gaussian target. Furthermore, varying the initial Task~0 class resulted in a final I-AUC standard deviation of only \textbf{0.21\%p}. These results demonstrate that DeCoFlow's effectiveness is invariant to the choice of the initial task, providing strong empirical support for the frozen-base approach.

\section{Conclusion}
\label{sec:conclusion}

We formalize subnet independence in affine coupling layers as a mechanism for parameter isolation in continual learning. DeCoFlow decomposes Normalizing Flows into a frozen base and task-specific low-rank adapters, structurally eliminating parameter-interference forgetting under correct routing, while TSA, ACL, and TAL compensate for frozen-base rigidity to deliver $FM=0.00\%$ on the evaluated task-agnostic streams with state-of-the-art detection on MVTec-AD and VisA.

Beyond accuracy, our analysis shows that coupling-internal decomposition is uniquely effective in preventing manifold collapse, and that DCL and ACL play complementary roles---dominant density transformation versus statistical normalization---elucidating the mechanism behind the adapter modules.

Limitations remain. The VisA localization gap motivates higher-resolution/multi-scale density modeling and stronger representations, and a diagnostic $4{\times}4$ mixed-CIL test on VisA shows lower stability when multiple normal distributions are mixed from the first task (I-AUC 77.97\%, P-AP 21.43\%). Since end-to-end forgetting is ultimately mediated by routing, the guarantee weakens as prototypes overlap: under far more similar classes or much longer streams, routing rather than parameter isolation becomes the limiting factor. Strengthening routing in such regimes, broader protocols (mixed $N{\times}M$, task-/domain-incremental, large-scale, and cross-domain CAD), prototype routing cost, and extensions to other generative models remain future work. More broadly, leveraging architecture-specific structural invariance for parameter isolation offers a principled path toward forgetting-free continual learning beyond normalizing flows.

\paragraph{Acknowledgements.}
This work was supported by Institute of Information \& Communications Technology Planning \& Evaluation (IITP) grant funded by the Korea government (MSIT) (RS-2024-00460011, Climate and Environmental Data Platform for Enhancing Climate Technology Capabilities in the Anthropocene (CEDP)) and (RS2021-II211343, Artificial Intelligence Graduate School Program (Seoul National University)). This work was also supported by the National Research Foundation of Korea (NRF) grant funded by the Korea government (MSIT) (RS-2024-00407803, RS-2025-23523657). This work was supported by the Korea Planning \& Evaluation Institute of Industrial Technology (KEIT), funded by the Ministry of Trade, Industry and Energy (MOTIE), Korea (No. RS-2026-25509027, Development and Demonstration of an Ontology- and On-Device AI-Based Autonomous Operation Agent System for Semiconductor Wafer Manufacturing Processes). This work was also supported by the BK21 FOUR Program (Education and Research Center for Industrial Innovation Analytics) funded by the Ministry of Education, Korea (No. 4120240214912).

%
%
\bibliographystyle{splncs04}
\bibliography{Reference,additional_references}

@inproceedings{bergmann2019mvtec,
  title={MVTec AD--A Comprehensive Real-World Dataset for Unsupervised Anomaly Detection},
  author={Bergmann, Paul and Fauser, Michael and Sattlegger, David and Steger, Carsten},
  booktitle={Proceedings of the IEEE/CVF conference on computer vision and pattern recognition},
  pages={9592--9600},
  year={2019}
}

@inproceedings{zou2022spot,
  title={SPot-the-Difference Self-supervised Pre-training for Anomaly Detection and Segmentation},
  author={Zou, Yang and Jeong, Jongheon and Pemula, Latha and Zhang, Dongqing and Dabeer, Onkar},
  booktitle={European Conference on Computer Vision},
  pages={392--408},
  year={2022},
  organization={Springer}
}

@inproceedings{li2017lwf,
  title={Learning without forgetting},
  author={Li, Zhizhong and Hoiem, Derek},
  booktitle={European conference on computer vision},
  pages={614--629},
  year={2017},
  organization={Springer}
}

@inproceedings{dinh2016realnvp,
  title={Density estimation using Real-NVP},
  author={Dinh, Laurent and Sohl-Dickstein, Jascha and Bengio, Samy},
  booktitle={International Conference on Learning Representations},
  year={2017}
}

@inproceedings{roth2022patchcore,
  title={Towards Total Recall in Industrial Anomaly Detection},
  author={Roth, Karsten and Pemula, Latha and Zepeda, Joaquin and Sch{\"o}lkopf, Bernhard and Brox, Thomas and Gehler, Peter},
  booktitle={Proceedings of the IEEE/CVF Conference on Computer Vision and Pattern Recognition},
  pages={14318--14328},
  year={2022}
}

@inproceedings{gudovskiy2022cflow,
  title={CFLOW-AD: Real-Time Unsupervised Anomaly Detection with Localization via Conditional Normalizing Flows},
  author={Gudovskiy, Denis and Ishizaka, Shun and Kozuka, Kazuki},
  booktitle={Proceedings of the IEEE/CVF Winter Conference on Applications of Computer Vision},
  pages={98--107},
  year={2022}
}

@inproceedings{uniad,
  title={A Unified Model for Multi-class Anomaly Detection},
  author={You, Zhiyuan and Cui, Lei and Shen, Yujun and Yang, Kai and Lu, Xin and Zheng, Yu and Le, Xinyi},
  booktitle={Advances in Neural Information Processing Systems},
  year={2022}
}

@article{fastflow,
  title={FastFlow: Unsupervised Anomaly Detection and Localization via 2D Normalizing Flows},
  author={Yu, Jiawei and Zheng, Ye and Wang, Xiang and Li, Wei and Wu, Yushuang and Zhao, Rui and Wu, Liwei},
  journal={arXiv preprint arXiv:2111.07677},
  year={2021}
}

@article{ewc,
  title={Overcoming catastrophic forgetting in neural networks},
  author={Kirkpatrick, James and Pascanu, Razvan and Rabinowitz, Neil and Veness, Joel and Desjardins, Guillaume and Rusu, Andrei A and Milan, Kieran and Quan, John and Ramalho, Tiago and Grabska-Barwinska, Agnieszka and others},
  journal={Proceedings of the national academy of sciences},
  volume={114},
  number={13},
  pages={3521--3526},
  year={2017},
  publisher={National Acad Sciences}
}

@inproceedings{mas,
  title={Memory Aware Synapses: Learning What (Not) to Forget},
  author={Aljundi, Rahaf and Babiloni, Francesca and Elhoseiny, Mohamed and Rohrbach, Marcus and Tuytelaars, Tinne},
  booktitle={European Conference on Computer Vision (ECCV)},
  pages={139--154},
  year={2018}
}

@inproceedings{gem,
  title={Gradient Episodic Memory for Continual Learning},
  author={Lopez-Paz, David and Ranzato, Marc'Aurelio},
  booktitle={Advances in Neural Information Processing Systems},
  year={2017}
}

@inproceedings{lora,
  title={LoRA: Low-Rank Adaptation of Large Language Models},
  author={Hu, Edward J and Shen, Yelong and Wallis, Phillip and Allen-Zhu, Zeyuan and Li, Yuanzhi and Wang, Shean and Wang, Lu and Chen, Weizhu},
  booktitle={International Conference on Learning Representations},
  year={2022}
}

@article{cadic,
  title={{CADIC}: Continual Anomaly Detection Based on Incremental Coreset},
  author={Yang, Gen and Deng, Zhipeng and Man, Junfeng},
  journal={arXiv preprint arXiv:2511.08634},
  year={2025}
}

@article{replaycad,
  title={ReplayCAD: Generative Diffusion Replay for Continual Anomaly Detection},
  author={Hu, Lei and Gan, Zhiyong and Deng, Ling and Liang, Jinglin and Liang, Lingyu and Huang, Shuangping and Chen, Tianshui},
  journal={arXiv preprint arXiv:2505.06603},
  year={2025}
}

@inproceedings{dne,
  title={Towards Continual Adaptation in Industrial Anomaly Detection},
  author={Li, Cheng-Lin and Lin, Tsung-Han and Tsai, Yung-Yu and Lin, Yen-Chang and Yang, Ming-Hsuan and Kira, Zsolt},
  booktitle={Proceedings of the 30th ACM International Conference on Multimedia},
  year={2022}
}

@inproceedings{surprisenet,
  title={Look At Me, No Replay! {SurpriseNet}: Anomaly Detection Inspired Class Incremental Learning},
  author={Lee, Anton and Zhang, Yaqian and Gomes, Heitor Murilo and Bifet, Albert and Pfahringer, Bernhard},
  booktitle={International Conference on Information and Knowledge Management},
  year={2023}
}

@inproceedings{ucad,
  title={Unsupervised Continual Anomaly Detection with Contrastively-learned Prompt},
  author={Liu, Jiaqi and Wang, Kai and Jiang, Qiang and Feng, Zhizhong and Zhang, Wei and You, Zheng},
  booktitle={Proceedings of the AAAI Conference on Artificial Intelligence},
  volume={38},
  year={2024}
}

@inproceedings{omnial,
  title={{OmniAL}: A Unified CNN Framework for Unsupervised Anomaly Localization},
  author={Zhao, Ying and Zhu, Kechen and Shi, Haoran and Guo, Yuqi and Bai, Chen},
  booktitle={Proceedings of the IEEE/CVF Conference on Computer Vision and Pattern Recognition},
  pages={3924--3933},
  year={2023}
}

@article{msflow,
  title={{MSFlow}: Multiscale Flow-Based Framework for Unsupervised Anomaly Detection},
  author={Zhou, Yixuan and Xu, Xing and Song, Jingkuan and Shen, Fumin and Shen, Heng Tao},
  journal={IEEE Transactions on Neural Networks and Learning Systems},
  year={2024},
  doi={10.1109/TNNLS.2023.3346541}
}

@inproceedings{hgad,
  title={Hierarchical Gaussian Mixture Normalizing Flow Modeling for Unified Anomaly Detection},
  author={Yao, Xincheng and Li, Ruoqi and Qian, Zefeng and Wang, Lu and Zhang, Chongyang},
  booktitle={European Conference on Computer Vision (ECCV)},
  pages={93--110},
  year={2024},
  organization={Springer}
}

@article{vqflow,
  title={{VQ-Flow}: Taming Normalizing Flows for Multi-Class Anomaly Detection via Hierarchical Vector Quantization},
  author={Zhou, Yixuan and Xu, Xing and Song, Jingkuan and Shen, Fumin and Shen, Heng Tao},
  journal={arXiv preprint arXiv:2409.00942},
  year={2024}
}

@incollection{mccloskey1989,
  title={Catastrophic Interference in Connectionist Networks: The Sequential Learning Problem},
  author={McCloskey, Michael and Cohen, Neal J},
  booktitle={Psychology of Learning and Motivation},
  volume={24},
  pages={109--165},
  year={1989},
  publisher={Academic Press}
}

@article{french1999,
  title={Catastrophic forgetting in connectionist networks},
  author={French, Robert M},
  journal={Trends in Cognitive Sciences},
  volume={3},
  number={4},
  pages={128--135},
  year={1999},
  publisher={Elsevier}
}

@inproceedings{dgr,
  title={Continual Learning with Deep Generative Replay},
  author={Shin, Hanul and Lee, Jung Kwon and Kim, Jaehong and Kim, Jiwon},
  booktitle={Advances in Neural Information Processing Systems},
  pages={2990--2999},
  year={2017}
}

@inproceedings{gainlora,
  title={Gated Integration of Low-Rank Adaptation for Continual Learning of Large Language Models},
  author={Liang, Yan-Shuo and Chen, Jia-Rui and Li, Wu-Jun},
  booktitle={Advances in Neural Information Processing Systems},
  year={2025}
}

@inproceedings{mingle,
  title={{MINGLE}: Mixtures of Null-Space Gated Low-Rank Experts for Test-Time Continual Model Merging},
  author={Qiu, Zihuan and Xu, Yi and He, Chiyuan and Meng, Fanman and Xu, Linfeng and Wu, Qingbo and Li, Hongliang},
  booktitle={Advances in Neural Information Processing Systems},
  year={2025}
}

@inproceedings{anacp,
  title={{AnaCP}: Toward Upper-Bound Continual Learning via Analytic Contrastive Projection},
  author={Momeni, Saleh and Noroozi, Vahid and others},
  booktitle={Advances in Neural Information Processing Systems},
  year={2025}
}

@article{calora,
  title={{C-LoRA}: Continual Low-Rank Adaptation for Pre-trained Models},
  author={Wang, Xiequn and Zhang, Runming and Liang, Yan-Shuo and Li, Wu-Jun},
  journal={arXiv preprint arXiv:2502.17920},
  year={2025}
}

@inproceedings{coso,
  title={Continuous Subspace Optimization for Continual Learning},
  author={Cheng, Quan and Wan, Yuanyu and Wu, Lingyu and Hou, Chenping and Zhang, Lijun},
  booktitle={Advances in Neural Information Processing Systems},
  year={2025}
}

@article{cfrdc,
  title={Context-aware Feature Reconstruction for Class-incremental Anomaly Detection and Localization},
  author={Pang, Jingxuan and Li, Chunguang},
  journal={Neural Networks},
  volume={181},
  pages={106788},
  year={2025},
  doi={10.1016/j.neunet.2024.106788}
}

@article{mtrmb,
  title={Multimodal Task Representation Memory Bank vs. Catastrophic Forgetting in Anomaly Detection},
  author={Zhou, You and Li, Xun and Ding, Jian},
  journal={arXiv preprint arXiv:2502.06194},
  year={2025}
}

@inproceedings{serra2018hat,
  title={Overcoming Catastrophic Forgetting with Hard Attention to the Task},
  author={Serr{\`a}, Joan and Sur{\'i}s, D{\'i}dac and Miron, Marius and Karatzoglou, Alexandros},
  booktitle={International Conference on Machine Learning},
  pages={4548--4557},
  year={2018},
  organization={PMLR}
}

@inproceedings{mallya2018packnet,
  title={{PackNet}: Adding Multiple Tasks to a Single Network by Iterative Pruning},
  author={Mallya, Arun and Lazebnik, Svetlana},
  booktitle={Proceedings of the IEEE Conference on Computer Vision and Pattern Recognition},
  pages={7765--7773},
  year={2018}
}

@article{rusu2016progressive,
  title={Progressive Neural Networks},
  author={Rusu, Andrei A and Rabinowitz, Neil C and Desjardins, Guillaume and Soyer, Hubert and Kirkpatrick, James and Kavukcuoglu, Koray and Pascanu, Razvan and Hadsell, Raia},
  journal={arXiv preprint arXiv:1606.04671},
  year={2016}
}

@inproceedings{shrivastava2016ohem,
  title={Training Region-Based Object Detectors with Online Hard Example Mining},
  author={Shrivastava, Abhinav and Gupta, Abhinav and Girshick, Ross},
  booktitle={Proceedings of the IEEE Conference on Computer Vision and Pattern Recognition},
  pages={761--769},
  year={2016}
}

@inproceedings{lin2017focal,
  title={Focal Loss for Dense Object Detection},
  author={Lin, Tsung-Yi and Goyal, Priya and Girshick, Ross and He, Kaiming and Doll{\'a}r, Piotr},
  booktitle={Proceedings of the IEEE International Conference on Computer Vision},
  pages={2980--2988},
  year={2017}
}

@article{cfa2022,
  title={{CFA}: Coupled-hypersphere-based Feature Adaptation for Target-Oriented Anomaly Localization},
  author={Lee, Sungwook and Lee, Seunghyun and Song, Byung Cheol},
  journal={IEEE Access},
  volume={10},
  pages={78446--78454},
  year={2022},
  doi={10.1109/ACCESS.2022.3193699}
}

@inproceedings{simplenet2023,
  title={SimpleNet: A Simple Network for Image Anomaly Detection and Localization},
  author={Liu, Zhikang and Zhou, Yiming and Xu, Yuansheng and Wang, Zilei},
  booktitle={Proceedings of the IEEE/CVF Conference on Computer Vision and Pattern Recognition},
  pages={20402--20411},
  year={2023}
}

@inproceedings{rd4ad2022,
  title={Anomaly Detection via Reverse Distillation from One-Class Embedding},
  author={Deng, Hanqiu and Li, Xingyu},
  booktitle={Proceedings of the IEEE/CVF Conference on Computer Vision and Pattern Recognition},
  pages={9737--9746},
  year={2022}
}

@inproceedings{iuf2024,
  title={An Incremental Unified Framework for Small Defect Inspection},
  author={Tang, Jiaqi and Lu, Hao and Xu, Xiaogang and Wu, Ruizheng and Hu, Sixing and Zhang, Tong and Cheng, Tsz Wa and Ge, Ming and Chen, Ying-Cong and Tsung, Fugee},
  booktitle={European Conference on Computer Vision (ECCV)},
  pages={306--323},
  year={2024},
  organization={Springer}
}

@inproceedings{cdad2025,
  title={One-for-More: Continual Diffusion Model for Anomaly Detection},
  author={Li, Xiaofan and Tan, Xin and Chen, Zhuo and Zhang, Zhizhong and Zhang, Ruixin and Guo, Rizen and Jiang, Guannan and Chen, Yulong and Qu, Yanyun and Ma, Lizhuang and Xie, Yuan},
  booktitle={IEEE/CVF Conference on Computer Vision and Pattern Recognition (CVPR)},
  year={2025}
}

@article{delange2021clsurvey,
  title={A Continual Learning Survey: Defying Forgetting in Classification Tasks},
  author={De Lange, Matthias and Aljundi, Rahaf and Masana, Marc and
          Parisot, Sarah and Jia, Xu and Leonardis, Ale{\v{s}} and
          Slabaugh, Gregory and Tuytelaars, Tinne},
  journal={IEEE Transactions on Pattern Analysis and Machine Intelligence},
  volume={44},
  number={7},
  pages={3366--3385},
  year={2022},
  doi={10.1109/TPAMI.2021.3057446}
}
\end{document}